\documentclass[journal]{IEEEtai}

\usepackage[colorlinks,urlcolor=blue,linkcolor=blue,citecolor=blue]{hyperref}
\usepackage{color,array}
\usepackage{graphicx}
\usepackage{array}

\usepackage[T1]{fontenc}
\usepackage[utf8]{inputenc}  
\usepackage{lmodern}    

\begin{document}

\title{Large Language Models in Operations Research: Methods, Applications, and Challenges} 

\author{
Yang Wang$^{1}$, Kai Li$^{2,3}$%
\thanks{Emails: yangw77718@gmail.com, kai.li@ia.ac.cn}\\[1ex]
$^{1}$University of Chinese Academy of Sciences, Beijing, China\\
$^{2}$Institute of Automation, Chinese Academy of Sciences, Beijing, China\\
$^{3}$School of Artificial Intelligence, University of Chinese Academy of Sciences, Beijing, China
}

\maketitle

\begin{abstract}
Operations research (OR) is a core methodology that supports complex system decision-making, with broad applications in transportation, supply chain management, and production scheduling. However, traditional approaches that rely on expert-driven modeling and manual parameter tuning often struggle with large-scale, dynamic, and multi-constraint problems, limiting scalability and real-time applicability. Large language models (LLMs), with capabilities in semantic understanding, structured generation, and reasoning control, offers new opportunities to overcome these challenges. It can translate natural language problem descriptions into mathematical models or executable code, generate heuristics, evolve algorithms, and directly solve optimization tasks. This shifts the paradigm from human-driven processes to intelligent human–AI collaboration.
This paper systematically reviews progress in applying LLMs to OR, categorizing existing methods into three pathways: automatic modeling, auxiliary optimization, and direct solving. It also examines evaluation benchmarks and domain-specific applications, and highlights key challenges, including unstable semantic-to-structure mapping, fragmented research, limited generalization and interpretability, insufficient evaluation systems, and barriers to industrial deployment. Finally, it outlines potential research directions.
Overall, LLMs demonstrate strong potential to reshape the OR paradigm by enhancing interpretability, adaptability, and scalability, paving the way for next-generation intelligent optimization systems.
\end{abstract}


\section{Introduction}

\IEEEPARstart{A}{s} a core methodology for decision-making in complex systems, OR has been widely applied in domains such as transportation, supply chain management, and production scheduling. However, with the rapid growth of data, increasing environmental dynamism, and rising demand customization, the traditional optimization paradigm—reliant on expert-based modeling and manual parameter tuning—has increasingly shown its limitations. The exponential expansion of problem scale, the growing complexity of constraint structures, and stringent real-time requirements pose formidable challenges to existing methods, significantly hindering their applicability and adoption in complex scenarios~\cite{1,2,3}.

Against this backdrop, the emergence of LLM offers a new opportunity to transform the paradigms of OR. Leveraging capabilities in semantic understanding, structured generation, and reasoning control, LLM has the potential to reshape the optimization process. Its value is primarily reflected in two aspects: First, at the modeling level, LLM can automatically translate natural language problem descriptions into formal mathematical models or executable code, thereby achieving a direct mapping from requirements to models~\cite{4,5}. Second, at the solving level, LLM can serve as a generator of heuristic strategies and collaborate with traditional optimization algorithms to build a closed-loop perception–reasoning–feedback framework, thereby improving both efficiency and solution quality~\cite{6,7,8}.

Although recent studies have made progress, most explorations remain confined to local breakthroughs on specific tasks, and a unified methodological framework has yet to emerge. When addressing complex logical reasoning, LLM still exhibits instability and uncertainty in results. The limited interpretability of its reasoning process further challenges the reliability of practical applications. Hence, a systematic review is needed to clarify the research landscape, unify methodological paradigms, and outline future development directions. As shown in Fig.~\ref{literature_classification}, the literature classification map visually illustrates the distribution of research pathways and representative works.
Specifically, this paper systematically introduces research on LLM-driven OR from three perspectives:

\begin{figure*}[t]
  \centering
  \includegraphics[width=0.8\linewidth]{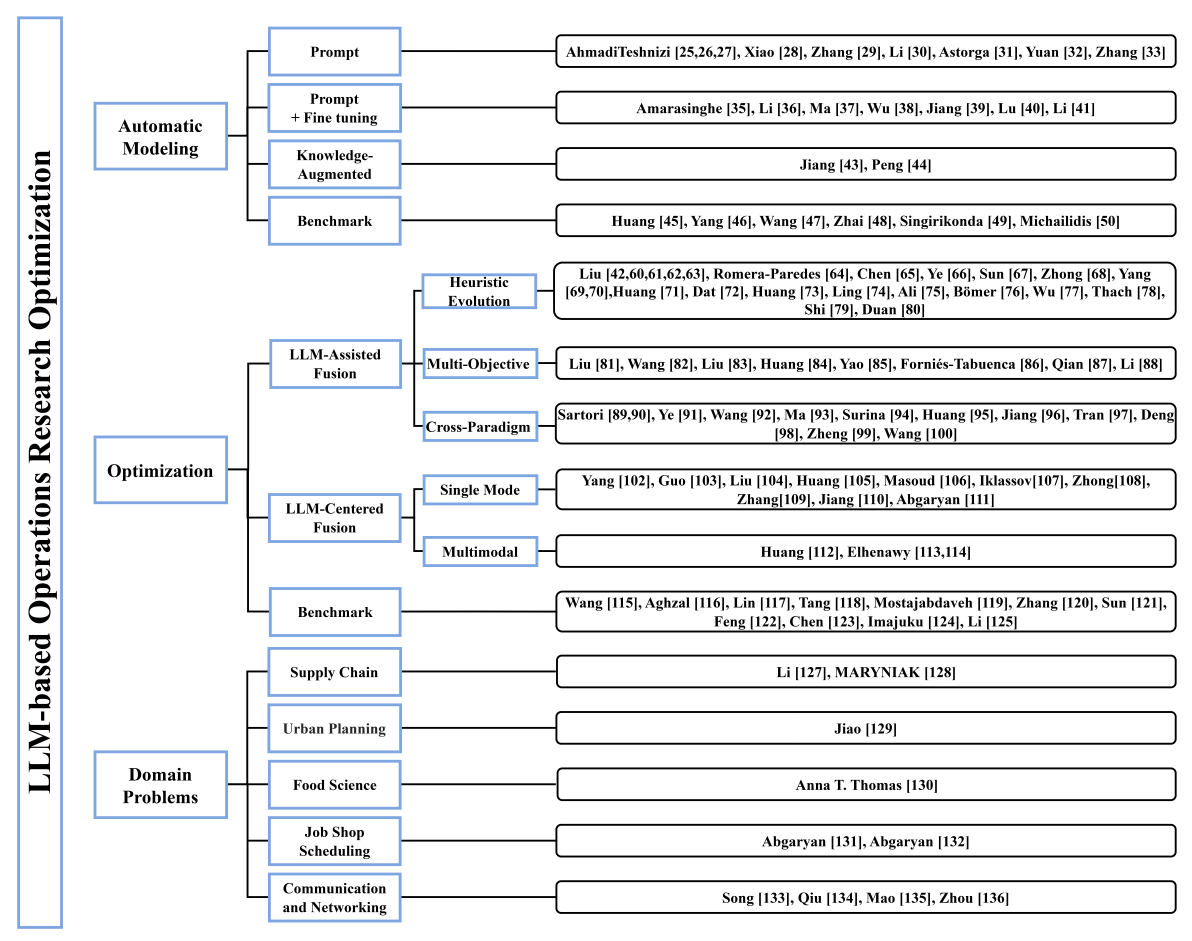}
  \caption{Literature classification map of LLM-driven OR research pathways and representative works.}
  \label{literature_classification}
\end{figure*}

\begin{itemize}
    \item \textbf{Methodological Paradigm Summary:} reviewing existing work by classifying it into modeling and solving paradigms, and summarizing core technologies and implementation paths.
    \item \textbf{Domain Application Analysis:} presenting case studies of LLM in typical scenarios such as supply chain optimization and urban management.  
    \item \textbf{Frontier Trends Outlook:} identifying key bottlenecks in current research and outlining potential breakthroughs and future directions.
\end{itemize}

Existing surveys on LLMs for optimization provide valuable but fragmented perspectives. Da Ros et al.~\cite{1} emphasize methodological advances within combinatorial optimization, whereas cross-paradigm and application-oriented perspectives remain underexplored. Liu et al.~\cite{2} highlight algorithm design, viewing LLMs mainly as generators rather than covering the full workflow. Xiao et al.~\cite{9} review the modeling stack—data, inference, and evaluation—yet give little attention to auxiliary optimization and solving. Zhang et al.~\cite{10} connect modeling and solving within evolutionary optimization but do not extend beyond this paradigm. In contrast, our review unifies automatic modeling, auxiliary optimization, and direct solving into a coherent framework, while also extending the analysis to domain problems and sector-specific applications, thereby offering a more comprehensive perspective on LLM-driven OR. The goal is to provide a systematic methodological review and structured insights, thereby promoting the deep integration of LLM with OR and supporting the development of next-generation intelligent decision-making systems.

\section{Introduction to Basic Principles}
\subsection{Operations Research (OR)}
OR is a systematic decision science centered on mathematical methods, aiming to obtain optimal or near-optimal solutions to objective functions under multiple constraints~\cite{11,12}. By means of formal modeling, complex real-world decision problems are abstracted into mathematical representations of objectives and constraints, providing the theoretical foundation and tools for systematic analysis and exact solving. The scope of OR covers diverse problem types, including linear programming, integer programming, and combinatorial optimization~\cite{13,14}. These models effectively capture structural characteristics and sources of uncertainty, thereby establishing the methodological basis for modeling and solving complex systems.
With its rigorous mathematical framework and broad applicability, OR has been widely adopted in critical domains such as transportation, supply chain management, and production scheduling~\cite{15,16}. However, as problem sizes expand and constraint structures become increasingly complex, traditional methods reveal clear limitations, including over-reliance on expert experience, insufficient modeling flexibility, and prohibitive computational costs for NP-hard problems. These challenges open the door for the introduction of LLM, which shows breakthrough potential in natural language–driven modeling, heuristic strategy generation, and the solving of complex problems.

\subsection{Foundation of Competence of LLM}
Pre-trained on large-scale corpora, LLM acquires semantic understanding, knowledge association, and generative reasoning~\cite{17,18}. Its relevance to OR lies in mapping natural language inputs into structured elements, algorithmic steps, or approximate solutions. Two core applications emerge: modeling, where LLM transforms unstructured descriptions into variables, objectives, and constraints to generate mathematical models or executable code~\cite{4,5,19}; and solving, where reasoning decomposition and chain-of-thought (CoT) support local search, heuristic generation, and tool invocation~\cite{20,21}. Moreover, LLM enables task generalization and interactive correction by suggesting parameters, producing heuristic operators, or reorganizing solving logic within optimization frameworks~\cite{6,7,22}. When modeling fails or solving degrades, it establishes a generate–validate–repair closed loop through multi-turn prompts, rewriting, or tool calls~\cite{23}, enhancing stability and controllability with self-debugging~\cite{21} and self-feedback~\cite{24}. These capabilities underpin the application of LLM in OR, particularly along two key pathways—automatic modeling and assisted solving—reviewed in the next section.

\section{LLM-based Methods for Operations Research}

\subsection{Automatic Modeling}

In recent years, automatic modeling has emerged as a core research direction for LLMs in OR. Ramamonjison et al.~\cite{4} first proposed a systematic modeling framework for translating natural language into optimization models, laying the research foundation for this field. Subsequently, Fan et al.~\cite{5} reviewed the multi-stage pathway for integrating artificial intelligence and OR, explicitly highlighting the unique potential of LLMs in the modeling phase. Huang et al.~\cite{19} further introduced the MAMO benchmark and proposed a “model generation + numerical validation” evaluation method, which for the first time established modeling capability as an independent research objective. As a result, automatic modeling has been increasingly recognized as a critical bridge between natural language and formal models~\cite{22}.

\begin{figure}[t]
  \centering
  \includegraphics[width=0.45\linewidth]{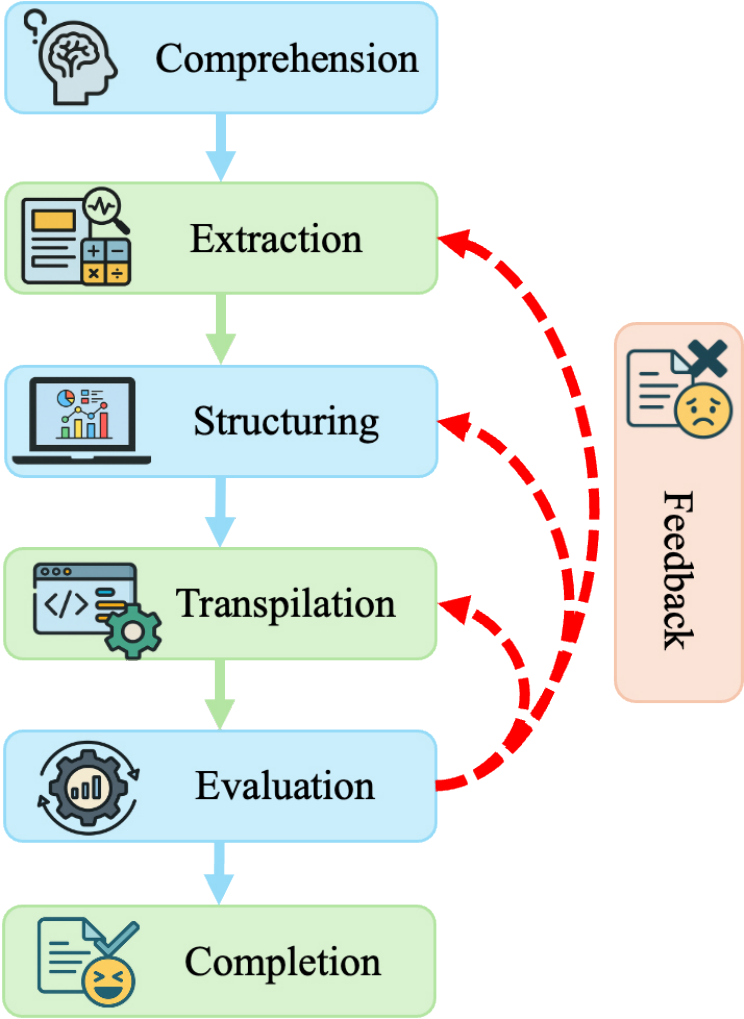}
  \caption{Closed-loop framework of automatic modeling from natural language input to model execution.}
  \label{auto_modeling_pipeline}
\end{figure}

To clarify the process, this paper summarizes automatic modeling into five sequential steps forming a closed loop from natural language to model execution, as illustrated in Fig.~\ref{auto_modeling_pipeline}. The process begins with problem comprehension, where optimization objectives and resource constraints are parsed, followed by element identification to extract decision variables, constraint types, and objective functions. These elements are then organized through structure generation into a standard mathematical model, which is further transformed into solver-executable code. Finally, verification and feedback are performed by running the code in a solver and iteratively correcting errors based on the results.

This section systematically reviews three representative research paradigms: (1) the pathway via prompting, (2) the collaborative pathway combining prompt and model fine-tuning, and (3) the external knowledge-guided mechanism. It also summarizes related evaluation benchmarks to compare their performance and applicability.

\subsubsection{LLM-based Automatic Modeling Paradigm via prompting}
The prompt-driven approach has become a significant direction for automated optimization modeling due to its lightweight design and ease of implementation. This class of methods guides the LLM to generate core components such as variables, objective functions, and constraints through prompting. Representative studies are summarized in Table~\ref{tab:prompt_only}.

\begin{table}[t]
\caption{Representative Studies on Prompt-based LLM Automatic Modeling}
\label{tab:prompt_only}
\tablefont
\footnotesize
\setlength{\tabcolsep}{2.5pt}
\renewcommand{\arraystretch}{1.15}

\begin{tabular}{|m{7pc}<{\centering}|
                m{9pc}<{\centering}|
                m{4pc}<{\centering}|}
\hline
\textbf{Framework} & \textbf{Key Modeling Traits} & \textbf{Benchmarks} \\
\hline
OptiMUS~\cite{25,26,27} & Multi-stage + Visualization + Reflection--retry prompts & NL4OPT + ComplexOR \\
Chain-of-Experts~\cite{28} & Multi-agent chain + Expert prompts + Backward reflection & LPWP + ComplexOR \\
OptLLM~\cite{29} & Three-stage dialogue + Interactive feedback & NL4OPT + Optimize tasks \\
NL2OR~\cite{30} & DSL generation + Structural validation & 30 OR inst. \\
Autoformulator~\cite{31} & Multi-stage + Composite prompts + Structural search & NL4OPT + IndustryOR + MAMO + ComplexOR \\
MA-GTS~\cite{32} & Multi-agent + Hierarchical modules + Semantic decomposition & G-REAL \\
OR-LLM-Agent~\cite{33} & Prompt modeling + Code validation + Self-repair chain + Validation chain & 83 real OR \\
\hline
\end{tabular}
\end{table}

In this pathway, the OptiMUS series of work by AhmadiTeshnizi et al. has built the most representative prompt-driven modeling process. The initial version~\cite{25} progressively converted natural language into a solvable optimization model, covering structure generation, code generation, execution validation, and feedback correction, and developed the accompanying NLP4LP dataset to support evaluation. Subsequent research~\cite{26} introduced a connection graph mechanism to record dependencies between variables, constraints, and parameters, thereby extracting context to prompt input and adapt to complex modeling scenarios. The latest version~\cite{27} further integrates a structure detection agent and a structure pool, combined with error correction mechanisms such as prompt retries and self-correction. It also supports calling solver subroutines in advanced optimization systems and provides an interactive modeling and visualization platform, thereby greatly improving the structural perception and engineering potential of the system.

Within prompting-based modeling, verification has become a central theme, though different works emphasize it from distinct angles. Xiao et al.~\cite{28} strengthen controllability through the Chain-of-Experts framework, where role-specific agents handle terminology parsing, model construction, code generation, and verification, supported by a two-stage reasoning mechanism of forward construction and backward reflection. Zhang et al.~\cite{29} highlight interactive refinement, employing a three-stage workflow of parsing, structural transformation, and result generation that integrates user feedback to lower the entry barrier for non-experts. Building on structured prompts, Li et al.~\cite{30} enhance structural correctness by introducing grammar correction, variable-consistency checks, and JSON Schema validation, with automatic restarts when errors occur. Astorga et al.~\cite{31} pursue systematic exploration, combining hierarchical decomposition and Monte Carlo Tree Search (MCTS) with structured templates and a candidate generation–pruning–ranking procedure to ensure both consistency and diversity.

Prompting has also been applied in specific scenarios. Yuan et al.~\cite{32} proposed the MA-GTS framework for graph-structured tasks, where multi-agent collaboration supports hierarchical modeling through semantic parsing, knowledge integration, and algorithmic solving, progressively reconstructing graph structures from text and invoking optimization algorithms adaptively. Zhang et al.~\cite{33} introduced the OR-LLM-Agent framework, which emphasizes reasoning-driven closed-loop modeling by using structured prompts to transform natural language into linear programming models, automatically generating executable code and performing repair and validation within a sandbox environment.

In summary, prompting has developed across multiple dimensions, encompassing both single-prompt guidance and multi-agent collaboration, incorporating interactive parsing and structured verification mechanisms, and spanning a wide range of scenarios from general tasks to domain-specific applications. These studies not only validate the effectiveness of prompting in automatic modeling but also highlight its potential in structural perception, task decomposition, and interactive generation.

\subsubsection{LLM-based Automatic Modeling Mechanism Driven by Prompt–Fine-tuning Synergy}
In research on automatic modeling using LLM, the synergistic mechanism of prompting and model fine-tuning has gradually emerged as a core developmental path. By combining input optimization with parameter updating, this mechanism alleviates the instability and limited generalization of single-prompt guidance in complex tasks, while further improving the model’s accuracy and robustness in variable recognition, constraint parsing, and structured representation~\cite{34}. A summary of representative studies is provided in Table~\ref{tab:prompt_finetune}.

\begin{table}[t]
\caption{Representative Studies on Prompt--Fine-tuning Synergy in Automatic Modeling}
\label{tab:prompt_finetune}
\tablefont
\footnotesize
\setlength{\tabcolsep}{2.5pt}      
\renewcommand{\arraystretch}{1.15} 

\begin{tabular}{|m{5pc}<{\centering}|
                m{10.5pc}<{\centering}|
                m{4.5pc}<{\centering}|}
\hline
\textbf{Study} & \textbf{Prompt Mechanism} & \textbf{FT} \\
\hline
Amarasinghe~\cite{35}  & Fixed structured prompts & SFT \\
Li~\cite{36}          & Template-based prompts & SFT \\
Ma~\cite{37}          & Structured prompts + Rewrite & CL + InstrSFT \\
Wu~\cite{38}           & Structured prompts + Validation chain & LoRA + Alpaca + COPT \\
Jiang~\cite{39}       & Structured prompts + Multi-rule aug. & MI-SFT + KTO \\
Lu~\cite{40}           & Structured prompts + Self-correction & LoRA \\
Li~\cite{41}           & Structure-enhanced NL prompts & SFT + DPO \\
\hline
\end{tabular}
\end{table}

In workflow and module design, researchers have explored decomposing complex tasks through structured prompts and staged fine-tuning. Amarasinghe et al.~\cite{35} proposed the AI Copilot framework, which uses supervised fine-tuning to transform natural language into modeling code and applies prompt engineering with nine submodules to overcome token limitations, ensuring completeness and executability. Li et al.~\cite{36} introduced a three-stage framework involving variable identification, constraint classification, and supplementary generation, which significantly improves the precision of variable extraction and constraint construction. Ma et al.~\cite{37} developed the LLaMoCo framework, incorporating code-to-code instruction tuning and a large-scale instruction set for diverse optimization tasks, while enhancing semantic alignment and constraint comprehension through unified structural prompts and diversified rewriting strategies. By integrating contrastive learning with instruction tuning in a two-stage process, LLaMoCo achieves stronger semantic understanding and generalization in code generation.

In the area of data generation and consistency verification, Wu et al.~\cite{38} proposed the Evo-Step-Instruct framework, which introduces dual strategies of complexity evolution and scope evolution to guide the model in generating diverse, high-quality data. A step-by-step verification mechanism is further employed to ensure consistency among descriptions, variables, and constraints, thereby effectively preventing error propagation. This method demonstrates strong stability and accuracy on complex benchmark tasks, underscoring the critical role of data quality in the effectiveness of synergistic mechanisms.

Meanwhile, modeling standardization and robustness enhancement have emerged as another important direction. Jiang et al.~\cite{39} proposed the LLMOPT framework, which defines a unified five-element modeling structure consisting of sets, parameters, variables, objectives, and constraints. By combining diverse samples generated through prompt templates with expert-annotated data, the framework employs supervised fine-tuning, model alignment, and self-correction mechanisms to significantly improve modeling reliability and solution stability in complex tasks. In contrast, Lu et al.~\cite{40} introduced the OptMATH framework, which emphasizes the construction of high-quality triplet datasets encompassing natural language, mathematical expressions, and solver code, while adjusting task difficulty through a feedback mechanism.

In the domain of structure-guided optimization algorithm generation, Li et al.~\cite{41} proposed the STRCMP framework, which integrates language modeling with graph structure learning. By extracting structural features through a constraint–variable bipartite graph, the framework generates solver-oriented algorithmic code conditioned jointly on natural language and structural embeddings. Leveraging supervised fine-tuning, preference optimization, and the generative–evolutionary capabilities of LLM, it iteratively optimizes performance across multiple rounds. Methodologically, it shares similarities with MA-GTS~\cite{32} in structural extraction and multi-module collaboration, while following the structure-guided code generation paradigm introduced by AEL~\cite{42}.

In summary, the synergy between prompting and fine-tuning demonstrates multi-dimensional advantages across workflow design, data generation, modeling standardization, and structure-guided algorithm generation. At the methodological level, it advances the systematic mapping from natural language to structured models, while at the practical level, it provides a solid foundation for scalability and industrial deployment. This synergistic mechanism lays the groundwork for extending automatic modeling to more complex tasks and cross-domain applications in the future.

\subsubsection{LLM-based Automatic Modeling Mechanism Guided by External Knowledge}

In complex modeling tasks, external knowledge guidance provides LLM with new avenues for enhancement. Jiang et al.~\cite{43} proposed the DRoC framework, which decomposes vehicle routing problems into constraint subtasks and injects precise external knowledge through semantic retrieval and document filtering.Moreover, the Bootstrap mechanism in DRoC further validates the feasibility of dynamically expanding external knowledge bases. Extending this direction, Peng et al.~\cite{44} focused on privacy-preserving and local deployment scenarios, proposing a domain knowledge–enhanced automatic modeling framework for constructing MILPs in multi-robot task allocation and scheduling problems. By guiding variable and constraint generation through knowledge bases, and combining prompt-based guidance with supervised fine-tuning, their method enables automatic generation of solver code, demonstrating strong stability and generative capability in representative scheduling tasks.

External knowledge guidance substantially enhances LLM’s modeling capacity for complex constraints and scheduling, improving accuracy and robustness while enabling dynamic expansion and scenario adaptation, thus supporting the development of more generalizable automatic modeling frameworks.

\subsubsection{Benchmarks for Evaluating LLM-based Automatic Modeling}

As the evaluation system for automatic modeling continues to mature, researchers have proposed more targeted benchmarks and frameworks from various perspectives, including industry coverage, dataset scale, structural equivalence, cross-task unification, and constraint programming adaptation. These efforts not only address the limitations of existing evaluation tools in terms of applicability and breadth but also drive the systematic assessment of automatic modeling capabilities across multiple dimensions.

In terms of industry adaptation and large-scale sample requirements, Huang et al.~\cite{45} proposed the OR-INSTRUCT framework and constructed the IndustryOR benchmark, which encompasses 1,556 natural language modeling problems across 16 industries, yielding a training set of 32,481 samples. The ORLM series models outperformed GPT-4 on the NL4OPT, MAMO, and IndustryOR tasks, demonstrating near-expert, human-level modeling capabilities under the Pass@8 setting. To address the limitations of small models in language understanding, Yang et al.~\cite{46} introduced the OptiBench and ReSocratic frameworks, constructing the ReSocratic-29K dataset. Through a reverse-generation strategy, they back-translated natural language problems and programs from structured modeling examples, significantly boosting the performance of the LLaMA series on MILP tasks. Recognizing the limited scope and insufficient domain coverage of existing benchmarks, Wang et al.~\cite{47} further extended OptiBench by proposing an evaluation tool comprising 816 problems across more than 80 domains, thereby strengthening cross-task and cross-domain coverage.

In structural equivalence, Zhai et al.~\cite{48} introduced EquivaFormulation on the NLP4LP dataset and developed the EquivaMap framework, which uses LLM to generate variable mappings and automatically assess semantic equivalence of modeling results. This addresses the limitations of traditional accuracy metrics and WL tests in complex scenarios. For cross-task evaluation, Singirikonda et al.~\cite{49} created the TEXT2ZINC dataset, standardizing on MiniZinc and covering 110 problems across 11 domains, including both optimization and satisfiability tasks. Results show that CoT reasoning and compositional modeling outperform basic prompting in generation accuracy, offering a robust benchmark for cross-paradigm evaluation.

In the direction of constraint programming, Michailidis et al.~\cite{50} proposed CP-Bench, which covers 101 combinatorial optimization problems and 241 types of constraints. The study systematically compared three modeling frameworks—MiniZinc, CPMpy, and OR-Tools—and introduced enhancement strategies including prompt design, in-context examples, repeated sampling, and self-verification. The results show that Python-based frameworks are more suitable for LLM-driven modeling, and that repeated sampling combined with self-verification can significantly improve solution accuracy.

In summary, these benchmarks and frameworks have expanded the evaluation dimensions of automatic modeling in terms of industry adaptation, dataset scale, structural validation, and cross-task assessment, thereby significantly advancing the systematic development of evaluation systems. Nevertheless, their limitations remain evident: task coverage is still largely confined to typical problems such as MILP and VRP, making it insufficient to capture complex constraints and heterogeneous modeling scenarios; evaluation metrics are overly centered on structural equivalence and accuracy, lacking systematic measures of efficiency, interpretability, and robustness; and some datasets rely on manual or synthetic generation, resulting in a substantial gap from real-world industrial requirements. Future research must therefore focus on developing more representative and practical benchmarks to support the evaluation of LLM’s modeling capabilities in diverse and realistic settings.

\subsection{LLM-assisted Optimization}

\begin{figure}[t]
  \centering
  \includegraphics[width=0.95\linewidth]{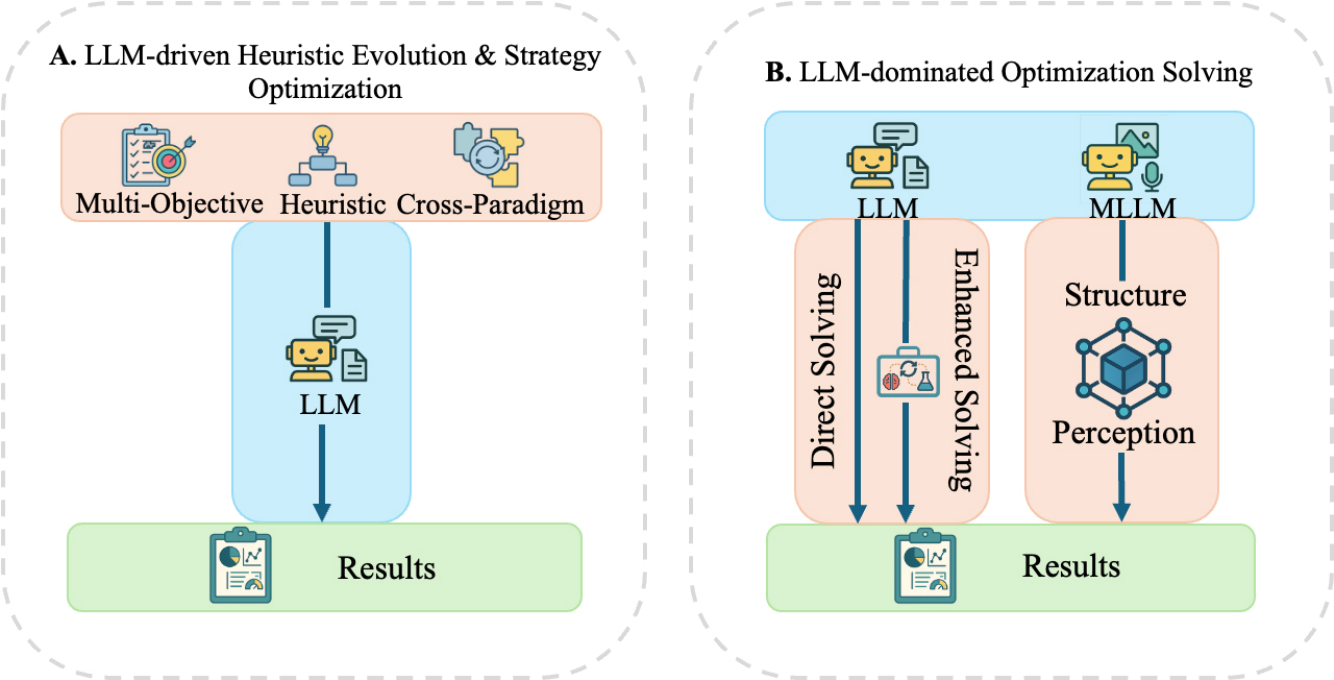}
  \caption{Two primary approaches to LLM-assisted optimization.}
  \label{llm_optimization}
\end{figure}

In recent years, LLM-assisted optimization has gradually become an important research direction in the field of OR. Its core objective is to enhance the practicality and controllability of models in solving optimization problems through structural design and capability coordination~\cite{22}. Figure~\ref{llm_optimization} illustrates the two primary approaches to LLM-assisted optimization.

This section systematically reviews two representative research directions: (1) LLM-enabled hybrid mechanisms for integrating multiple optimization algorithms, and (2) LLM-dominated optimization solving. It also summarizes and analyzes the related evaluation benchmarks to comprehensively present the performance and limitations of such methods in optimization tasks.

\subsubsection{LLM-enabled Hybrid Mechanisms for Integrating Multiple Optimization Algorithms}

In recent years, LLM has demonstrated unique advantages in reasoning control and strategy generation, opening new avenues for their integration with diverse optimization algorithms. Existing studies show that LLM can contribute not only to operator design and search guidance but also to strategy adjustment and structural optimization. Huang et al.~\cite{51}, through an analysis of structural mapping between LLM and evolutionary algorithms, revealed their synergistic potential in operator generation and search control; Cai et al.~\cite{7} further proposed leveraging LLM as a component for structural guidance and strategy adjustment, effectively enhancing the automation and adaptability of the optimization process. Against this backdrop, this paper further reviews three representative fusion pathways: (1) heuristic structure evolution and strategy optimization; (2) collaborative mechanisms in multi-objective evolution; and (3) capability transfer in cross-paradigm integration.

LLM has demonstrated broad potential in its integration with evolutionary algorithms. Existing studies have examined diverse embedding methods and collaborative pathways, encompassing heuristic structure generation, strategy regulation, and the optimization of evolutionary mechanisms~\cite{52,53}, while emphasizing LLM’s critical role in search guidance and strategy automation~\cite{54,55}. This direction has further extended to the joint evolution of objective functions and control modules~\cite{56,57}, revealing the behavioral patterns and performance evolution of LLM in heuristic strategy development~\cite{58,59}. Overall, this approach has established a systematic framework, with the core mechanisms of related studies summarized in Table~\ref{tab:llm-evolution}.

\begin{table}[t]
\caption{Representative Studies on LLM-driven Heuristic Structure Evolution and Strategy Optimization}
\label{tab:llm-evolution}
\tablefont
\footnotesize
\setlength{\tabcolsep}{3pt}
\renewcommand{\arraystretch}{1.15}

\begin{tabular}{|m{6pc}<{\centering}|m{14pc}<{\centering}|}
\hline
\textbf{Study} & \textbf{Core Mechanism} \\
\hline
Liu~\cite{60} & Zero-shot operator + Temperature adaptation + Natural language \\
Liu~\cite{42,61,62} & Structured prompt modeling + Multi-round evolutionary optimization \\
Liu \& Li~\cite{63} & Constraint-aware heuristic construction for VRP \\
Romera-Paredes~\cite{64} & LLM program generation with evaluator, maintain search diversity \\
Chen~\cite{65} & QUTC design, UIQ-based parent selection \\
Ye~\cite{66} & Short- and long-term reflection for evolution \\
Sun~\cite{67} & Module replacement + Candidate evaluation \\
Zhong~\cite{68}& Multi-module prompt-based heuristic generation \\
Yang~\cite{69,70} & Multi-agent strategy evolution and selection \\
Huang~\cite{71} & Reflection + Scheduling rule evolution \\
Dat~\cite{72} & Role prompts + Flash reflection + Harmony \\
Huang~\cite{73} & Reasoning path generation + Modular strategies \\
Ling~\cite{74} & “Explore” + “Modify” strategy \\
Ali~\cite{75} & Human preference-based selection \\
Bömer~\cite{76}& Semantic structure + Context-driven generation \\
Wu~\cite{77} & Key component extraction + Performance prediction \\
Thach~\cite{78} & Language reduction + Parallel evolution \\
Shi~\cite{79}& Dual-loop mechanism \\
Duan~\cite{80} & Instance generator + Solver co-evolution \\
\hline
\end{tabular}
\end{table}

On direct embedding strategies, Liu et al.~\cite{60} proposed the LMEA framework, which employs LLMs as zero-shot operators for parent selection, crossover, and mutation. This approach significantly reduces the complexity of operator design and domain adaptation. Candidate solutions are generated through natural language prompts, while a temperature-adaptive mechanism balances exploration and exploitation, yielding superior performance over traditional methods on multiple TSP instances.

Building on this idea, Liu et al.~\cite{42} introduced the AEL framework, which regards the optimization algorithm itself as the target. By using structured prompts, LLMs generate, rewrite, and refine algorithms, enabling iterative self-improvement across evolutionary rounds. Further extensions~\cite{61} applied AEL to guided local search, where LLMs automatically produce key guiding functions that outperform manually designed schemes, demonstrating strong generalization on complex TSP tasks.

Continuing along this trajectory, the authors developed the LLM4AD platform~\cite{62}. It leverages structured prompts and search mechanisms to generate standardized, executable algorithmic code, supported by a unified evaluation pipeline that forms a complete closed loop. In domain-specific applications, Li et al.~\cite{63} proposed the ARS framework, which uses prompt engineering to construct constraint-aware heuristics for vehicle routing problems. The framework is supported by the RoutBench benchmark, which covers 1,000 VRP variants and six real-world constraints, and achieves end-to-end optimization by performing constraint selection, detection, and scoring function generation without fine-tuning.

On algorithm generation and evolutionary mechanisms, Romera-Paredes et al.~\cite{64} proposed the FunSearch framework, which employs LLM as a core generator to iteratively produce, evaluate, and retain high-quality solutions within the function space, while maintaining search diversity through a distributed architecture. Building on this, Chen et al.~\cite{65} introduced the QUBE framework, which incorporates a unified uncertainty-based indicator to guide parent selection and population resetting, thereby achieving a dynamic balance between structural exploration and local exploitation. Ye et al.~\cite{66} proposed the ReEvo framework, embedding a dual-layer language reflection mechanism into the heuristic evolutionary loop. After generating heuristic code, the model can suggest improvements and guide subsequent searches, enabling strategies to progressively converge toward superior solution spaces.

On solver module optimization, Sun et al.~\cite{67} proposed the AutoSAT framework, which decomposes CDCL solvers into modular components and leverages LLM to generate candidate functions to replace specific modules. High-quality alternatives are retained through performance evaluation, thereby progressively enhancing solver performance. On structured prompting strategies, Zhong et al.~\cite{68} introduced the CRISPE prompting strategy, composed of five submodules, to guide LLM in generating heuristic pseudocode. Based on this, they developed the metaheuristic algorithm ZSO, which demonstrated strong performance and convergence stability across CEC2014, CEC2022, and several engineering optimization problems.

Within multi-agent collaboration, Yang et al.~\cite{69} proposed the HeurAgenix framework, which constructs a heuristic optimization system composed of four agents—generation, evolution, evaluation, and selection—thereby extending the role of LLM from a single generator to a multi-role collaborator. Further research~\cite{70} introduced a two-stage hyper-heuristic framework that integrates solution trajectories with preference mechanisms to iteratively refine heuristic strategies, even surpassing traditional specialized optimizers.

For dynamic scheduling and diversity maintenance, Huang et al.~\cite{71} proposed the SeEvo framework, which treats LLM-generated scheduling rules as individuals in the population and employs both individual and collective reflection mechanisms for continuous optimization, outperforming traditional methods in dynamic scheduling tasks. Dat et al.~\cite{72} introduced the HSEvo framework, which leverages role-playing prompts to generate diverse heuristic individuals and integrates Harmony Search for local refinement, effectively enhancing adaptability while maintaining diversity.

In graph optimization and complex planning scenarios, Huang et al.~\cite{73} proposed the GraphThought framework, which leverages reasoning-path generation and template synthesis mechanisms to advance structural recognition and LLM-based strategy integration in graph optimization tasks. Ling et al.~\cite{74} introduced the AutoHD framework, which employs LLM to generate diverse heuristic functions and continuously evolve them, thereby improving solution quality and reasoning efficiency in complex planning tasks.

In the field of preference modeling and task adaptability, Ali et al.~\cite{75} proposed the PAIR framework, which incorporates human-like preference mechanisms and leverages structured prompts to perform high-quality individual pairing and crossover and mutation, thereby equipping LLM with selection and regulation capabilities in heuristic evolution. Bömer et al.~\cite{76} introduced the CEoH framework, which integrates task context with structured prompts to generate heuristic algorithms that are more targeted and adaptive. Wu et al.~\cite{77} proposed the Hercules framework, which introduces performance prediction and confidence control mechanisms to achieve a coordinated trade-off between quality and efficiency during heuristic generation and evaluation.

In problem reformulation and meta-optimization exploration, Thach et al.~\cite{78} proposed the RedAHD framework, which employs language-based simplification to automatically generate surrogate problems and conducts multi-source evolution across multiple spaces, thereby extending the boundaries of heuristic discovery. Shi et al.~\cite{79} introduced the MoH framework, which adopts a bi-level optimization process: in the outer loop, LLM generates candidate optimizers, while in the inner loop, the optimizers generate heuristics. This approach overcomes the limitations of fixed optimizer structures and demonstrates strong generalization ability in multi-task environments. Duan et al.~\cite{80} proposed the EALG framework, which establishes an adversarial co-evolutionary system in which LLM generates problem instances and heuristic solvers separately. Driven by adversarial objectives, both evolve jointly, enhancing problem difficulty and strategy adaptability.

In summary, LLM in heuristic structure evolution and strategy optimization has progressed from a single generator to a full-process collaborator, advancing heuristic design from experience-driven patterns toward adaptive evolution. Building upon this foundation, research has further extended into multi-objective evolutionary optimization, where the collaborative role of LLM evolves from offspring generation to the construction of systematic frameworks, as outlined in Table~\ref{tab:moe-llm-collab}.

\begin{table}[t]
\caption{LLM Collaboration in Multi-objective Evolution}
\label{tab:moe-llm-collab}
\tablefont
\footnotesize
\setlength{\tabcolsep}{3pt}
\renewcommand{\arraystretch}{1.12}

\begin{tabular}{|m{4pc}<{\centering}|m{16pc}<{\centering}|}
\hline
\textbf{Study} & \textbf{Key Idea} \\
\hline
\multicolumn{2}{|c|}{\textbf{Phase I: Offspring Generation and Search Collaboration}} \\
\hline
Liu~\cite{81} & Prompted offspring + Linear operator to reduce cost while preserving generalization \\
Wang~\cite{82} & Embed LLM as search operator in CCMO + Co-evolve with classic genetic operators \\
Liu~\cite{83} & Low-cost adaptive MOEA with auxiliary evaluation + Stagnation detection + NSGA-II \\
\hline
\multicolumn{2}{|c|}{\textbf{Phase II: Evolutionary Mechanism Construction}} \\
\hline
Huang~\cite{84} & LLM-generated executable mutation operators + Performance feedback \\
Yao~\cite{85} & Evolve non-dominated heuristic sets with dominance + Code diversity + Guided parents + Population update \\
Forniés-Tabuenca~\cite{86} & NSGA-II guided parents + Clustering-based reflection for heuristic evolution \\
\hline
\multicolumn{2}{|c|}{\textbf{Phase III: Unified System Framework}} \\
\hline
Qian~\cite{87} & Unified framework + Homogeneous crossover + Heterogeneous co-evolution + Upgrades \\
\hline
\multicolumn{2}{|c|}{\textbf{Task-specific Applications}} \\
\hline
Li~\cite{88} & Portfolio MO with NSGA-II + Non-dominated sorting + Crowding distance validation \\
\hline
\end{tabular}
\end{table}

Early studies primarily explored the auxiliary role of LLM in offspring generation and search operators. Liu et al.~\cite{81} were the first to introduce LLM into multi-objective evolutionary optimization, where the MOEA/D-LLM framework employed prompt engineering to generate offspring individuals and designed linear operators to reduce invocation costs while maintaining strong generalization performance. Wang et al.~\cite{82} embedded LLM as a novel search operator within the CCMO framework, guiding the generation of partial offspring solutions and co-evolving with traditional genetic operators, thereby accelerating convergence and improving solution quality. Subsequently, Liu et al.~\cite{83} proposed a low-cost adaptive mechanism that invokes LLM to generate candidate solutions only when evolutionary stagnation occurs, and integrates them with NSGA-II operators, thereby improving solution quality under limited computational resources.

With the growing generative capabilities of LLM, its role has gradually expanded to the construction of evolutionary operators and heuristic structures. Huang et al.~\cite{84} proposed an automatic operator generation framework that guides LLM to produce executable mutation operators from structured task descriptions and iteratively refines them based on performance feedback, thereby enhancing algorithm adaptability and structural flexibility. Yao et al.~\cite{85} introduced the MEoH framework, which employs a combined strategy of dominance and structural diversity to guide parent selection and population updating, automatically evolving a structurally diverse set of non-dominated heuristics. Forniés-Tabuenca et al.~\cite{86} developed the REMoH framework, which represents individuals using LLM-generated heuristic functions and incorporates NSGA-II’s non-dominated sorting and crowding-distance mechanisms, while embedding a clustering-based reflection process to improve solution diversity and robustness.

Building on mechanism exploration, research has further advanced toward systematic integration. Qian et al.~\cite{87} proposed the MLHH framework, which unifies the evolution of heuristic structures through three steps: homogeneous crossover evolution, heterogeneous co-evolution, and architecture function upgrading. The framework executes a “generate–standardize–evaluate–select” loop iteratively, enabling the continuous optimization of high-quality multi-objective solution strategies. This stage of work highlights the shift in LLM’s role in multi-objective evolution from supporting localized operations to orchestrating holistic system-level coordination.

For real-world tasks, Li et al.~\cite{88} embedded LLM into the NSGA-II framework and proposed the llmPC-NSGA-II method for multi-objective portfolio optimization. This approach generates complete offspring populations through structured prompts and updates them using non-dominated sorting and crowding distance mechanisms. Experimental results on real financial data and standard benchmark functions demonstrate strong convergence and diversity, validating the application potential of LLM in practical OR problems.

Overall, the collaborative mechanisms of LLM in multi-objective optimization follow a clear trajectory from initial offspring generation to unified frameworks and real-world applications, positioning LLM as a transformative force in this domain. Moving beyond the scope of multi-objective problems, recent studies have begun to emphasize cross-paradigm integration, where LLM’s language understanding, structural generation, and reasoning abilities are combined with classical operators, reinforcement learning, and neuro-symbolic systems to enhance transferability and global optimization capacity. To systematically present the progress of cross-paradigm integration, Table~\ref{tab:llm-fusion-landscape} summarizes representative studies and their primary technical pathways.

\begin{table}[t]
\caption{Representative Studies on LLM Capability Transfer and Synergy in Cross-paradigm Fusion}
\label{tab:llm-fusion-landscape}
\tablefont
\footnotesize
\setlength{\tabcolsep}{3pt}
\renewcommand{\arraystretch}{1.12}

\begin{tabular}{|m{3pc}<{\centering}|m{17pc}<{\centering}|}
\hline
\textbf{Study} & \textbf{Key Idea} \\
\hline
\multicolumn{2}{|c|}{\textbf{Fusion with Classical Optimization}} \\
\hline
Sartori~\cite{89} & Prompted node features + Probabilistic BRKGA decoder + Structure-aware metaheuristics \\
Sartori~\cite{90} & Heuristic context + Policy redesign + Preference correction-penalty \\
Ye~\cite{91} & Bi-level evolution + NL/Code heuristics + Memory-guided evolution \\
Wang~\cite{92} & SA with heuristic reconstruction and temperature adaptation \\
\hline
\multicolumn{2}{|c|}{\textbf{Fusion with Reinforcement Learning}} \\
\hline
Ma~\cite{93} & Heuristic pool + RL policy selection + Feedback-driven improvement \\
Surina~\cite{94} & Program sampling + DPO-LLM optimization + Preference construction \\
Huang~\cite{95} & Prompted structure + GRPO-RL optimization + Collapse-restart evolution \\
\hline
\multicolumn{2}{|c|}{\textbf{Fusion with neuro-symbolic systems}} \\
\hline
Jiang~\cite{96} & Semantic encoding + RL-trained generator + Instance understanding \\
Tran~\cite{97} & LLM attention bias + POMO/LEHD integration + Lightweight fine-tuning \\
\hline
\multicolumn{2}{|c|}{\textbf{Search Trajectory Control and Reasoning}} \\
\hline
Deng~\cite{98} & Spatial prompting + Q-learning correction + Reverse curriculum learning \\
Zheng~\cite{99} & Heuristic actions + Dual-call mechanism + Thought-aligned search \\
Wang~\cite{100} & MCTS framework + Trajectory evaluation + State–action–reward chain \\
\hline
\end{tabular}
\end{table}

Some studies have focused on coupling LLM with classical optimization operators, embedding it into traditional algorithms such as large neighborhood search (LNS) and simulated annealing to enhance search efficiency and structural adaptability. Sartori et al. proposed an LLM-based heuristic redesign approach in two studies: on the one hand, structured prompts were used to extract node features and generate probabilistic signals, which were embedded into a BRKGA decoder to achieve structure-aware metaheuristic search~\cite{89}; on the other hand, the contextual understanding capabilities of LLM were incorporated into CMSA to reconstruct heuristic functions, generating variants with age bias and entropy regularization, thereby improving structural diversity and solution quality~\cite{90}. Similarly, Ye et al.~\cite{91} designed a bi-level evolutionary LNS, where the inner layer leverages LLM to generate heuristic strategies, while the outer layer evolves prompt templates to enhance diversity. The framework also introduces differential memory and adaptive perturbation mechanisms to balance search efficiency and convergence control. Meanwhile, Wang et al.~\cite{92} integrated LLM into simulated annealing, employing three-stage prompts to guide new solution generation and combining them with a temperature control mechanism to enhance the ability to escape local optima.

Beyond traditional operators, some studies have explored the integration of LLM with RL to support strategy optimization and adaptive evolution. Ma et al.~\cite{93} proposed the AutoDH framework, which constructs a pool of heuristic functions and employs an RL agent to select the optimal function based on solution states, thereby enabling subpath optimization and feedback-driven improvement. Surina et al.~\cite{94} introduced the Evo-Tune framework, which samples high-quality candidate construction prompts during program search to guide LLM-generated programs. After performance verification, these candidates form a preference dataset used to update the language model through direct preference optimization, gradually biasing it toward generating higher-quality structures. Building on this, Huang et al.~\cite{95} proposed the CALM framework, which incorporates prompt diversification and collapse-restart mechanisms, combined with generalized reinforcement preference optimization, to continuously enhance structural generation capabilities, thereby improving stability and robustness in complex tasks.

At a higher level of exploration, researchers have sought to integrate LLM into neuro-symbolic systems to unify semantic representation and structural constraints. Jiang et al.~\cite{96} proposed combining LLM with Transformers to map problem instances into a unified semantic space, and employing RL to train the generator, thereby improving solution quality and diversity. Tran et al.~\cite{97} developed a neural optimizer framework in which LLM automatically generates structure-aware attention biases to guide node selection. This mechanism was integrated into multi-objective optimization and lightweight enhanced heuristic decoders, achieving cross-scale generalization under lightweight fine-tuning conditions.

In addition, some studies have focused on search trajectory control and reasoning mechanisms. Deng et al.~\cite{98} proposed the S2RCQL model for path planning tasks, which transforms spatial prompts into entity relationships to guide LLM in constructing path representations, and integrates Q-learning with reverse curriculum learning to reduce learning difficulty, thereby improving reasoning stability and generalization. Zheng et al.~\cite{99} introduced the MCTS-AHD framework, where LLM generates heuristic functions and semantic descriptions within MCTS, and a two-stage invocation aligns function code with conceptual descriptions to enhance semantic consistency and interpretability. In contrast, Wang et al.~\cite{100} proposed the PoH framework, which employs MCTS as the dominant component while invoking LLM at the trajectory and key-node levels for evaluation and refinement, forming a “state–action–reward” closed-loop decision process that further improves global search performance.

Overall, cross-paradigm integration has opened new directions for the application of LLM in optimization algorithms. From coupling with traditional operators, to deep collaboration with RL, and further to extensions into neuro-symbolic systems and reasoning mechanisms, the transfer and coordination of LLM capabilities is gradually forming a systematic pathway.

\subsubsection{LLM-dominated Optimization Solving}

In the evolution of LLM-assisted OR, research that directly leverages model generation to solve optimization problems has gradually become a frontier direction. These approaches no longer rely on traditional mathematical modeling or heuristic algorithms. Instead, they employ natural language or multimodal prompts to guide LLM in generatively addressing diverse optimization tasks~\cite{101}. To systematically review their development trajectory and application characteristics, this paper classifies related studies into two categories: single-modal generative optimization methods and multimodal structure-aware methods, which will be further analyzed through representative works in the following sections.

Single-modal generative optimization methods leverage natural language or structured text prompts to directly guide LLM in producing optimization solutions, thereby eliminating the dependence on gradient information and explicit modeling processes. These methods demonstrate advantages such as generality, interactive flexibility, and ease of transfer. Related studies have undergone an evolutionary process from direct solving to technique-enhanced approaches, and further to unified solving of complex constraints. Their core mechanisms and application tasks are summarized in Table~\ref{tab:llm-direct-solve}.

\begin{table}[t]
\caption{Representative Studies on Single-modal Generative Optimization Methods}
\label{tab:llm-direct-solve}
\tablefont
\footnotesize
\setlength{\tabcolsep}{3pt}
\renewcommand{\arraystretch}{1.12}

\begin{tabular}{|m{4pc}<{\centering}|m{16pc}<{\centering}|}
\hline
\textbf{Study} & \textbf{Key Mechanism} \\
\hline
\multicolumn{2}{|c|}{\textbf{Phase I: Direct Solving by LLM}} \\
\hline
Yang~\cite{102} & NL prompting + Meta-prompt iteration \\
\hline
\multicolumn{2}{|c|}{\textbf{Phase II: Fusion-driven Techniques}} \\
\hline
Guo~\cite{103} & CoT + Historical prompts + Interactive optimization \\
Liu~\cite{104} & State prediction + Utility modeling + Preference ranking  \\
Huang~\cite{105} & NL-to-Python + Self-debugging and verification \\
Masoud~\cite{106} & Prompt-driven + Multi-sampling + Self-ensemble \\
Iklassov~\cite{107} & Meta-prompt + Multi-trajectory + Recursive subtask optimization \\
Zhong~\cite{108} & Self-evolving prompt + Accuracy loop \\
Zhang~\cite{109} & SFT + RLF + RAG \\
Jiang~\cite{110} & Structural prompt + Multi-round opt. + Knowledge integration \\
\hline
\multicolumn{2}{|c|}{\textbf{Phase III: Exploring LLM’s Solving Capacity}} \\
\hline
Abgaryan~\cite{111} & Constraint-aware output + Dynamic control \\
\hline
\end{tabular}
\end{table}

Early studies primarily focused on direct solving, aiming to validate the black-box optimization capability of LLM under a “zero-modeling” setting. A representative example is the OPRO method proposed by Yang et al.~\cite{102}, which requires neither gradients nor heuristic operators but relies solely on natural language descriptions and historical solution feedback to construct meta-prompts that guide iterative solution generation by the LLM. In both continuous and combinatorial optimization tasks, OPRO demonstrated the potential for autonomous exploration of the solution space and achieved measurable improvements in accuracy in language reasoning tasks, indicating that LLM can perform basic optimization without external tool support. However, the limited performance of such methods in complex structures and multi-round tasks has driven research toward the incorporation of enhancement mechanisms.

Subsequent studies have progressively enhanced the solving capabilities of LLM in a gradual manner. Early improvements focused on reasoning support and trajectory interpretability, such as combining CoT reasoning with iterative prompting to accumulate the dialogue history as a traceable “optimization memory,” thereby enabling the transition from “one-shot generation” to “interactive convergence” under a unified evaluation framework~\cite{103}. On the other hand, researchers have investigated decision-making in uncertain environments. By incorporating state prediction, utility modeling, and preference ranking, LLM can simulate the process of expected utility maximization and adaptively balance gains and risks across multiple rounds of feedback, demonstrating greater robustness in real-world domains such as agriculture and finance~\cite{104}. In addition, the work of Huang et al.~\cite{105} achieved end-to-end closed-loop solving by transforming natural language descriptions into executable code and integrating self-debugging and self-verification mechanisms in the feedback loop to ensure the feasibility and stability of generated solutions. In vehicle routing problem experiments, this method demonstrated that even without manual modeling or traditional heuristic support, LLM possesses strong zero-shot optimization capabilities.

Beyond closed-loop design, studies have also sought to improve solution diversity and stability. Masoud et al.~\cite{106} proposed a self-ensemble strategy that mitigates local convergence through multi-round sampling and selection, enabling cross-scale generalization in tasks such as the TSP. Other work explored trajectory generation and recursive decomposition, where meta-prompts produce multiple strategies that are decomposed into subtasks and recursively optimized, extending solution scope and diversity across combinatorial, scheduling, and symbolic reasoning tasks~\cite{107}. Zhong et al.~\cite{108} further introduced a self-evolving prompt mechanism driven by accuracy feedback, in which iterative “generate–evaluate–update” cycles refine prompt design, sustaining stable performance in high-dimensional structural search and supporting cross-scale generalization.

With research progress, structured input scenarios have attracted growing attention. Zhang et al.~\cite{109} built a large-scale graph–task–code dataset and introduced a two-stage training process with retrieval-augmented mechanisms, enabling LLM to map natural language graph tasks to executable code and achieve cross-task generalization. Jiang et al.~\cite{110} addressed the structural matrix ordering problem using prompts based on network topology, node descriptions, and semantic context to guide LLM in generating node sequences without gradients or operators, while applying multi-round iterative optimization to accelerate convergence and improve solution quality. These advances highlight the ability of single-modal generative approaches to handle higher-dimensional structural problems.

Researchers have increasingly focused on unified solving under complex constraints. Abgaryan et al.~\cite{111} proposed the ACCORD framework, which embeds constraint control into the output format to ensure satisfaction during decoding. Combined with task identification and LoRA routing, ACCORD enables unified solving across multiple NP-hard problem classes. This work highlights the potential of LLM in tackling optimization tasks with complex constraints and diverse problem types, offering a new perspective for general-purpose language-based optimizers.

In summary, single-modal methods have advanced from direct generation to reasoning- and feedback-enhanced solving, but remain limited in capturing complex structures and multi-source information. To address this gap, multimodal structure-aware approaches integrate visual and linguistic inputs to improve spatial perception and structural consistency, showing particular promise for tasks such as path planning and scheduling. The following section extends this discussion to other pathways, highlighting how external knowledge and hybrid mechanisms further enhance the modeling and solving capabilities of LLM in OR.

In structure-aware multimodal approaches, Huang et al.~\cite{112} proposed the MLLM-V framework, which integrates text and image prompts for solving the capacitated vehicle routing problem, simulating human cognitive processes through heuristic extraction, solution generation, and feedback refinement. This demonstrates the value of visual information in enhancing the quality and generalization of generative optimization. Building on this line, Elhenawy et al.~\cite{113} introduced a multimodal reasoning framework that combines image inputs with natural language prompts, enabling LLM to intuitively solve the traveling salesman problem (TSP) and progressively refine path structures through adaptive mechanisms, thereby highlighting the potential of multimodal inputs. Furthermore, Elhenawy et al.~\cite{114} proposed a purely vision-driven framework with both three-stage and dual-agent collaborative architectures, capable of generating solutions for TSP and multiple TSP without relying on coordinates or distance matrices, underscoring the promise of visual perception in complex structural optimization.

Overall, multimodal structure-aware approaches enhance LLM-based optimization by integrating visual and linguistic inputs, thereby improving structural understanding, stability, and generalization. Compared with single-modal methods, they offer complementary mechanisms that drive the evolution of optimization, and are expected to extend to broader inputs such as tabular and sensor data for more versatile systems.

\subsubsection{Benchmarks for Evaluating LLM-assisted Optimization}

To systematically evaluate the role of LLM in solving optimization problems, researchers have proposed a series of evaluation benchmarks. Unlike traditional tests that primarily assess modeling outcomes, these benchmarks place greater emphasis on reasoning and decision-making capabilities within complex optimization processes, covering a wide range of tasks such as graph-structured reasoning, path planning, task generation, optimization interpretation, and strategy improvement.

For graph reasoning, Wang et al.~\cite{115} proposed the NLGraph benchmark, which contains 29,370 problems spanning eight categories of graph reasoning tasks, providing an important tool for solving graph problems in natural language settings. For path planning in spatiotemporal reasoning, Aghzal et al.~\cite{116} constructed the PPNL benchmark, which simulates grid-world path planning scenarios to systematically evaluate model performance across varying prompt strategies and task complexities. Focusing on plan reasoning in asynchronous multi-step tasks, Lin et al.~\cite{117} introduced the AsyncHow benchmark, which includes 1,600 high-quality instances across multiple common domains such as education, diet, health, and household activities.

In graph computation tasks, Tang et al.~\cite{118} proposed the GraphArena benchmark, which employs a three-stage evaluation process consisting of path extraction, feasibility verification, and optimality checking. It covers four classes of polynomial-time tasks and six classes of NP-complete problems. This benchmark not only provides detailed distinctions between correct, suboptimal, hallucinatory, and missing outputs but also enhances the interpretability and diagnostic capability of the reasoning process. The authors further explored four enhancement strategies—CoT prompting, instruction fine-tuning, code generation, and reasoning augmentation—examining their applicability limits and potential utility in complex graph problems.

For the systematic evaluation of reasoning capabilities in OR, Mostajabdaveh et al.~\cite{119} proposed the ORQA benchmark, which consists of 1,513 multiple-choice questions spanning 20 application domains in OR. To analyze the importance of interpretability in optimization processes, Zhang et al.~\cite{120} constructed the first industrial-scale dataset for explainable optimization, covering 30 problem types and 300 queries, and introduced the EOR framework. This framework incorporates the concept of “decision information” and combines what-if analysis with graph-structured modeling. Through the collaboration of three agent modules—Commander, Writer, and Safeguard—it generates dual outputs of code modification explanations and result variation explanations, offering a new pathway to enhance the interpretability of LLM in optimization tasks.

For evaluating algorithm generation and end-to-end task execution in combinatorial optimization, Sun et al.~\cite{121} proposed CO-Bench, which encompasses 36 real-world problem classes, including bin packing, cutting, location, path planning, and tree-structure optimization. The benchmark provides natural language descriptions, data-loading functions, and evaluation functions, forming a complete workflow spanning data reading, algorithm design, and performance evaluation. Building on this, Feng et al.~\cite{122} developed the FrontierCO benchmark, which systematically evaluates an LLM solver across eight combinatorial optimization tasks. Results indicate that the LLM solver is generally more robust and exhibits stronger generalization capability compared to neural approaches, and in certain tasks, it rivals or even surpasses state-of-the-art heuristic methods. However, its performance fluctuates significantly across tasks, revealing shortcomings in algorithm selection and integration scheduling.

To validate the intelligence of LLM in heuristic design, Chen et al.~\cite{123} proposed the HeuriGym framework, which employs an iterative mechanism of “prompt generation–code execution–error feedback–multi-round refinement.” The framework establishes a benchmark covering nine real-world tasks and introduces the Quality-Yield Index as a core metric, comprehensively evaluating the reasoning ability of the intelligent agent through success rate and relative solution quality. Focusing on long-term goal-driven structural optimization, Imajuku et al.~\cite{124} developed ALE-Bench, constructed on the AtCoder Heuristic Contest, encompassing NP-hard problems such as path planning and production scheduling. It is equipped with scorers, visualization tools, and expert evaluation systems to assess LLM’s optimization ability under both single-round and multi-round iterations.

For large-scale search spaces, Li et al.~\cite{125} proposed OPT-BENCH, which covers 20 real-world machine learning tasks and 10 classical NP-hard combinatorial optimization problems. Accompanying it, the OPT-Agent framework simulates the human problem-solving process of “draft–optimize–debug,” supporting the entire pipeline from initial solution generation to stepwise refinement based on feedback, thereby forming a fully automated workflow that integrates task definitions, datasets, metrics, and validation scripts.

Overall, existing benchmarks provide essential tools for evaluating LLM in optimization, supporting their transition from modeling assistants to intelligent collaborators. Yet they remain limited in task diversity, evaluation dimensions, and realism, with most focusing on static combinatorial problems and narrow metrics. Future work should expand coverage to dynamic, multimodal, and cross-domain scenarios, while establishing unified, integrated benchmarks that capture efficiency, stability, and interpretability in real-world settings.

\section{Domain Problems}
With increasingly complex real-world application scenarios, the demand for optimization methods continues to grow, further exposing the limitations of traditional OR regarding modeling complexity, interactivity, and interpretability. The introduction of LLM offers new perspectives for addressing these challenges, not only enhancing problem analysis and knowledge representation but also demonstrating potential advantages in interactive processes and result presentation~\cite{126}. Therefore, this section, together with the methodological pathways, systematically discusses the practical applications and future trends of LLM in typical scenarios.

\subsection{Supply Chain}
In supply chain optimization, LLM enhances interpretability by bridging natural language with solver execution. OptiGuide~\cite{127}, which integrates GPT-4 with an optimizer, exemplifies the prompt-driven modeling pathway: it translates user requirements into optimization expressions, forms a closed feedback loop, and achieves over 90\% accuracy in Microsoft Azure practice. Complementing such applications, a theoretical framework~\cite{128} systematically mapped LLM functions across supply chain stages, highlighting roles in supplier evaluation, forecasting, routing, and sustainability. Together, these studies demonstrate how LLM extends automatic modeling toward structured decision support and foreshadow external knowledge–guided mechanisms in OR.

\subsection{Urban Planning}
As urban systems continue to evolve, governance processes face increasingly complex challenges. While traditional OR methods possess strong theoretical rigor, their practical application in smart cities characterized by multi-source data and high-frequency responsiveness remains challenging. By combining reasoning and language understanding capabilities, LLM offers a novel pathway for complex decision-making in urban contexts. The City-LEO framework~\cite{129} integrates LLM reasoning with end-to-end optimization, interpreting users’ natural language requirements and leveraging historical data to generate relevant objective functions and reduce problem scale. At the same time, by combining random forests with mixed-integer programming, the framework enables interpretable mappings from features to decisions, producing solutions with higher computational efficiency and greater transparency. Experimental results show that the framework outperforms traditional methods in computational efficiency, global suboptimality control, and local satisfaction improvement, highlighting the synergistic effect of LLM and traditional optimization algorithms under a hybrid paradigm.
\subsection{Food Science}
In the domain of sustainable food management, traditional OR approaches are increasingly constrained by modeling complexity, high expertise requirements, and limited interpretability. Recent researcher~\cite{130} have proposed a framework that integrates the knowledge of LLM with human preference modeling capabilities, coupled with combinatorial optimization techniques, to construct a complete workflow that transforms natural language inputs into optimized decision-making outputs. Experimental evaluations demonstrate that this framework can substantially reduce greenhouse gas emissions while preserving overall user satisfaction, thereby underscoring the fusion of prompt-driven pathways with structural mapping mechanisms as a promising direction for intelligent optimization.
\subsection{Job Shop Scheduling}
The Job Shop Scheduling Problem (JSSP) is one of the most challenging tasks in OR, having long attracted attention due to its strong constraints and combinatorial complexity. In recent years, researchers have explored incorporating LLM into JSSP solution frameworks. One line of work constructed a large-scale dataset comprising 120,000 randomly generated natural language descriptions and feasible solutions, and employed LoRA to fine-tune the Phi-3-Mini model while introducing sampling strategies to enhance scheduling performance~\cite{131}. Subsequently, the authors introduced the Starjob dataset, extended the approach to the LLaMA model, and completed single-GPU training via Rank-Stabilized LoRA. On the Tai and DMU benchmarks, their method significantly outperformed traditional heuristic and neural network baselines~\cite{132}. Collectively, these explorations highlight the potential of integrating model fine-tuning mechanisms with structured task prompting, providing new ideas for intelligent solving of complex scheduling problems.
\subsection{Communication and Networking}
In mobile edge computing optimization, researchers have undertaken diverse explorations across related tasks. For server allocation, researchers have proposed a natural language interaction framework that leverages multi-round prompting and feedback to guide LLM in generating user–server assignment schemes that satisfy constraints while minimizing latency~\cite{133}. In the task of wireless access point deployment, the LMCO framework employs structured prompts to generate deployment schemes and incorporates ray-tracing evaluation and propagation knowledge to accelerate convergence~\cite{134}. For critical node identification, the problem has been reformulated as a score-function generation task, leading to the design of an LLM-driven evolutionary optimization framework for automatically identifying the most important nodes in communication networks~\cite{135}. At the system level, further studies have explored the potential of LLM in resource scheduling, prompt design, and heuristic strategy generation, embedding them into multiple optimization paradigms and thereby extending its application boundaries in intelligent communication systems~\cite{136}.

\section{Conclusion and Outlook}
With the continuous advancement of LLM in natural language understanding, structural generation, and reasoning control, its role in OR is shifting from exploratory studies to paradigm reconstruction and system integration. This paper reviewed three core research pathways—automatic modeling, assisted optimization, and direct solving. LLM has demonstrated capabilities in translating natural language into structured models, generating heuristics and strategies for complex optimization tasks, and enabling end-to-end problem solving.
Despite these advances, key challenges remain: unstable semantic–structure mapping; fragmented research outcomes without a unified framework; limited generalization and interpretability; insufficient evaluation tools and benchmarks; and high computational cost for large-scale deployment. Addressing these issues requires progress in robust representation and closed-loop mechanisms, standardization of task abstractions and workflows, integration of symbolic reasoning and causal analysis, development of multidimensional benchmarks, and exploration of lightweight deployment strategies.
Overall, research on LLM-enabled OR is entering a stage of systematic development. By advancing methodological innovation, enhancing interpretability, and establishing comprehensive evaluation systems, LLM is expected to drive the construction of next-generation intelligent optimization systems.

\vfill
\begin{flushleft}
\scriptsize
This work has been submitted to the IEEE for possible publication. 
Copyright may be transferred without notice, after which this version may no longer be accessible.
\end{flushleft}

\end{document}